\documentclass[lettersize,journal]{IEEEtran}
\usepackage{amsmath,amsfonts}
\usepackage{algorithmic}
\usepackage{algorithm}
\usepackage{array}
\usepackage[caption=false,font=normalsize,labelfont=sf,textfont=sf]{subfig}
\usepackage{textcomp}
\usepackage{stfloats}
\usepackage{url}
\usepackage{verbatim}
\usepackage{graphicx}
\usepackage{cite}
\usepackage{amsmath}  
\usepackage{amssymb}
\usepackage{amsthm}
\usepackage{booktabs}
\usepackage{algorithm}
\usepackage{algorithmic}
\usepackage{lineno}
\usepackage{arydshln}
\usepackage[normalem]{ulem}
\begin{document}

\title{PortraitTalk: Towards Robust and Customizable Audio-to-Talking Face Generation}

\author{Fatemeh Nazarieh, Zhenhua Feng*,~\IEEEmembership{Senior Member,~IEEE}, Diptesh Kanojia, Muhammad Awais, Josef Kittler,~\IEEEmembership{Life Member,~IEEE}
        % <-this % stops a space

%\thanks{Manuscript received May 12, 2025}
\thanks{F. Nazarieh, D. Kanojia, M. Awais, and J. Kittler are with the School of Computer Science and Electronic Engineering, University of Surrey, Guildford GU2 7XH, Surrey, UK. M. Awais and J. Kittler are also with the Centre for Vision, Speech and Signal Processing, University of Surrey.}
\thanks{Z. Feng is with the School of Artificial Intelligence and Computer Science, Jiangnan University, Wuxi 214122, Jiangsu, China.}
\thanks{*Corresponding Author: Zhenhua Feng}
}

% % The paper headers
% \markboth{Journal of \LaTeX\ Class Files,~Vol.~14, No.~8, August~2021}%
% {Shell \MakeLowercase{\textit{et al.}}: A Sample Article Using IEEEtran.cls for IEEE Journals}

%\IEEEpubid{0000--0000/00\$00.00~\copyright~2021 IEEE}
%\IEEEpubidadjcol
% Remember, if you use this you must call \IEEEpubidadjcol in the second
% column for its text to clear the IEEEpubid mark.

\maketitle

\begin{abstract}
    Recent advancements in audio-driven talking face generation have largely focused on achieving accurate audio-lip synchronization, often at the expense of visual realism, personalization, and generalizability, which are key factors in producing convincing talking face videos. To address these limitations, we present PortraitTalk, a robust and customizable audio-driven talking face generation framework designed for enhanced visual quality and user control. Our method is based on a latent diffusion architecture comprising two key modules: IdentityNet and AnimateNet. Specifically, IdentityNet preserves the identity across the generated video frames, and AnimateNet ensures smooth and coherent facial motion over time. This approach reduces reliance on reference-style videos prevalent in existing approaches. Another notable contribution of our work is the integration of text prompts through decoupled cross-attention mechanisms, enabling greater creative flexibility in video generation. Last, we propose a new evaluation metric, namely Audio-Driven Facial Dynamics (ADFD), that effectively measures the quality of a generated video in terms of both spatial and temporal aspects. Extensive experiments demonstrate that our method outperforms the state-of-the-art approaches, offering a robust and practical solution for realistic, customizable talking face synthesis driven by audio input. 
    % The source code of the proposed method is available at \url{https://github.com/armtp/PortraitTalk.git}.

    % The source code of the proposed method will be made publicly available at \url{https://github.com/armtp/PortraitTalk.git} upon acceptance.

\end{abstract}


\section{Introduction}
\IEEEPARstart{R}{ecently}, audio-to-talking face generation has gained significant attention within the generative AI community, primarily due to its broad spectrum of applications, including video content creation, animation production, video dubbing, etc. The aim of audio-to-talking face generation is to create realistic talking face videos precisely synchronized with the provided audio input~\cite{surveypaper,edtalk}. 
The studies in this area can be broadly divided into three categories: enhancing visual quality, improving audio–lip synchronization, and incorporating emotion into generated talking faces. While existing methods have achieved notable progress in each aspect, they often struggle to jointly preserve head pose control, expressive variation, identity consistency, and fine-grained facial detail customization. One common strategy to address these limitations is to decompose facial motion into separate components, such as mouth movements, head orientation, and eye blinks, and then integrate them during the rendering phase~\cite{9878472,Chen2019HierarchicalCT,makeittalk}. Although this modular approach can improve controllability, it is not without difficulties.
Many implementations rely heavily on predefined pose coefficients extracted from 3D face models~\cite{10814703,Tan_2024_CVPR,Peng_2024_CVPR,Xu_2024_CVPR,Peng_2023_ICCV}, which increases dataset requirements and can propagate modeling inaccuracies~\cite{edtalk}. 
Moreover, such systems typically require full retraining when adapting to a new identity, which limits their scalability in practical applications. 

In addition to these motion modeling constraints, most existing methods also lack flexibility in handling multiple input modalities for richer control. Current talking face generation frameworks are predominantly driven by either audio~\cite{makeittalk,Gan_2023_ICCV,wav2lip} or video~\cite{EAMM,Pang_2023_CVPR,Gong_2023_ICCV} inputs, offering limited adaptability in multimodal contexts. Recently, text-driven talking face generation has attracted growing interest~\cite{11045546,Song_2022_CVPR,li2021writeaspeaker,tan2024style2talker,Jang_2024_CVPR}. However, its potential for fine-grained customization remains largely underexplored. Existing approaches often employ textual descriptions in a narrow way, primarily for reconstructing or setting the scene, rather than enabling rich control over visual attributes. In practice, a more capable framework should harness textual input to manipulate a wide range of elements, from facial attributes and expressions to background composition, object presence, and overall visual style. Such flexibility is essential for producing versatile and realistic results in applications spanning virtual assistants, animated characters, and personalized digital content.
Beyond customization, practical deployment also demands the ability to generalize to identities unseen during training and to produce realistic results from only a few reference images. 
This capability removes the need for costly identity-specific retraining and extends applicability to real-world scenarios where visual data is scarce. 
Achieving this while supporting multimodal control remains challenging, requiring precise identity preservation, temporal coherence, and accurate audio–lip synchronization.

To mitigate the above issues, this paper presents PortraitTalk, a robust and customizable audio-to-talking face generation framework that integrates audio, visual, and textual modalities to enable fine-grained facial attribute control, coherent background generation, and accurate audio–lip synchronization for unseen identities. 
Specifically, PortraitTalk comprises two pivotal components: IdentityNet and AnimateNet. 

\textbf{IdentityNet} preserves consistent identity across the generated video frames, while integrating semantics from text descriptions into the generated faces. 
To enable customization, we integrate text and image embeddings in IdentityNet through the decoupled cross-attention mechanism~\cite{ye2023ipAdapter}. 
Traditional approaches typically concatenate text-image pairs in text-driven talking face generation~\cite{li2021writeaspeaker,ma2023dreamtalk} or audio-image pairs in audio-driven talking face generation~\cite{Shen_2023_CVPR,EAMM,10229247,stableTalk} within a singular attention block. 
We argue that this approach restricts a generative model from thoroughly learning and integrating identity-specific facial details and textual descriptions. 
To further enhance generation quality, we incorporate a mask reconstruction loss into IdentityNet. In this setup, a portion of input frames is randomly corrupted, and the model is trained to reconstruct the original frames. Inspired by masked language modeling~\cite{sinha2021masked} in natural language processing, this strategy encourages the model to infer missing visual information from the surrounding context and guiding conditions. The underlying hypothesis is that by learning to recover corrupted regions using unmasked areas, the network implicitly gains a stronger understanding of visual structure and facial coherence, leading to improved generation fidelity.
Through extensive experiments, we find that IdentityNet, enhanced with mask reconstruction and decoupled cross-attention, not only preserves high-fidelity identity across frames but also enables consistent talking-face generation without altering core facial features. By integrating textual semantic description within the latent space, it supports fine-grained editing, such as adjusting hairstyle, hair color, background, or environmental settings, all without the need for post-processing. 
 
Other critical quality factors in talking face generation are audio-lip synchronization and temporal consistency. Even minor artifacts or inconsistent head movements can noticeably reduce visual realism. To address this, \textbf{AnimateNet} goes beyond simply mapping audio to facial motion. It leverages a motion generator~\cite{ye2023geneface++} to extract facial motion representations from audio, forming the first phase of the animation pipeline. This stage improves the generalization capability of PortraitTalk to diverse speech beyond the training data. In the second phase, AnimateNet transforms these motion cues into realistic video frames through a diffusion-based architecture enhanced with dedicated cross-attention mechanisms. Specifically, it incorporates structure-aware attention for precise audio-lip synchronization, identity-guided attention for consistent head placement, and temporal attention for smooth transitions between frames. This design ensures accurate lip movements, consistent head placement, and smooth transitions, resulting in coherent and identity-preserving talking face videos.

Despite the notable progress in audio-to-talking face generation, the existing \textbf{evaluation metrics} often concentrate solely on isolated attributes such as visual quality or lip synchronization. 
Consequently, these metrics cannot provide a comprehensive assessment that reflects the holistic performance of a method. 
There exists an imperative need for a new metric that encompasses both spatial and temporal consistencies to facilitate thorough evaluation. 
Spatial consistency is crucial as it ensures the coherence and stability of the generated video frames, thereby preserving the integrity of identity and expression throughout the video. 
Temporal consistency, on the other hand, assesses the smoothness and natural flow of the transition over time, which are vital for producing lifelike and persuasive talking faces. 
In this paper, we propose a novel Audio-Driven Facial Dynamics (ADFD) measurement that addresses the above deficiencies by integrating both spatial and temporal factors, thereby providing a reliable measure of the overall quality of a generated talking face video. 

\begin{figure}[!t]
    \centering
    \includegraphics[trim={1.2cm 10.5cm 1.2cm 1.0cm}, clip, width=\columnwidth]{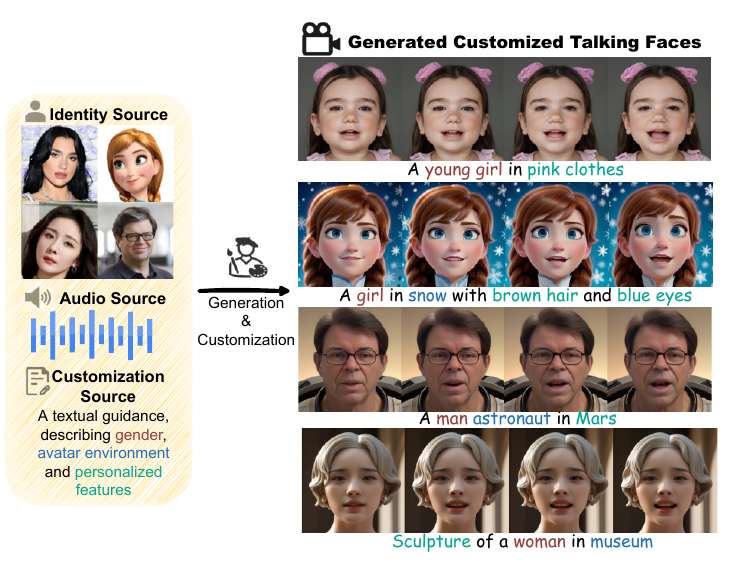}
    \caption{Talking faces generated by PortraitTalk. Given the reference images of an identity and the corresponding audio input, PortraitTalk synthesizes high-quality talking face videos that closely preserve the identity's appearance and speaking style. Furthermore, the visual attributes of the generated video, such as hair color, age, environmental settings and facial expressions, can be flexibly customized using a simple text prompt. This enables fine-grained control over the generated content, allowing users to tailor the visual presentation to specific narrative or stylistic requirements.
    }
    \label{fig0}
\end{figure}

By tackling the aforementioned challenges, PortraitTalk represents a significant step forward in audio-to-talking face generation, offering a more robust and flexible solution that enhances the realism and practicality of generated talking face videos. 
Some examples of the generated talking faces by PortraitTalk are shown in Fig.~\ref{fig0}.
Separating IdentityNet and AnimateNet allows each module to specialize in distinct aspects of the generation process. IdentityNet focuses on preserving the identity and reflecting text prompts, while AnimateNet is designed to handle motion-related information by mapping audio-driven motion priors and applying geometry control conditions to the vision domain. In summary, the main contributions of PortraitTalk include:
\begin{itemize}
    \item We develop a robust and customizable audio-to-talking face generation framework that can generalize well to unseen faces. We use three different modalities in the proposed framework.
    \item We propose a new evaluation metric that effectively measures video quality in terms of both spatial and temporal aspects.
    \item Qualitative and quantitative experiments demonstrate the effectiveness of PortraitTalk in enhancing synchronization, realism, and customization of generated talking faces as compared with the state-of-the-art methods.
\end{itemize}

The rest of this paper is organized as follows: We first introduce the existing literature in Section~\ref{Sec_rw}.
Then Section~\ref{sec_method} presents the proposed PortraitTalk method in detail. 
Last, the experimental results are reported and analyzed in Section~\ref{sec_exp}, and the conclusion is drawn in Section~\ref{conclusion}.

\section{Related Work}
\label{Sec_rw}
\subsection{Audio-driven Talking Face Generation}

In the field, various methods have been proposed, with intermediately guided talking face generation being particularly common.
Such models leverage intermediate representations, such as 2D facial landmarks or 3D faces, to effectively bridge the gap between audio input and visual output. 
These methods~\cite{makeittalk,10286358,9681173,Zhou2021Pose,Gan_2023_ICCV,9894719,9917325} typically consist of two main components: the first predicts intermediate representations from audio, and the second uses these features to generate talking faces. 
This two-stage methodology facilitates control over the synchronization between audio and visual data, thereby enhancing the realism and accuracy of the generated talking faces.

Chen et al. \cite{Chen2019HierarchicalCT} proposed a method that converts audio into mouth landmarks to generate talking faces. 
However, as it only predicts movements for the lower half of the face and uses the upper half from ground truth frames, the resulting videos often lack realism. 
MakeItTalk~\cite{makeittalk} introduced a one-shot audio-to-talking-face generation model based on facial landmarks. 
While it can handle various identities, MakeItTalk struggles with precise audio-lip synchronization and often fails to deliver high-quality talking faces. 
The Audio2Head~\cite{wang2021audio2head} model generates talking faces by extracting relatively dense motion fields from audio, but the faces often exhibit distortions and lack consistency in identity preservation.  
SadTalker~\cite{Zhang_2023_CVPR} has significantly enhanced the generalization capability through a learned latent space. Despite this advancement, the head movements, which are generated based on predefined motion coefficients, frequently lack realism. 
IP-LAP~\cite{Zhong_2023_CVPR}, a transformer-based generative model, utilizes audio inputs and sketches to produce talking faces. Although it achieves stable head movements, it falls short in synchronizing lip movements with audio, particularly with identities unseen during training.

\begin{figure*}[!t]
    \centering
    \includegraphics[trim={0cm 9.0cm 8.5cm 0cm}, clip, width=.95\textwidth]{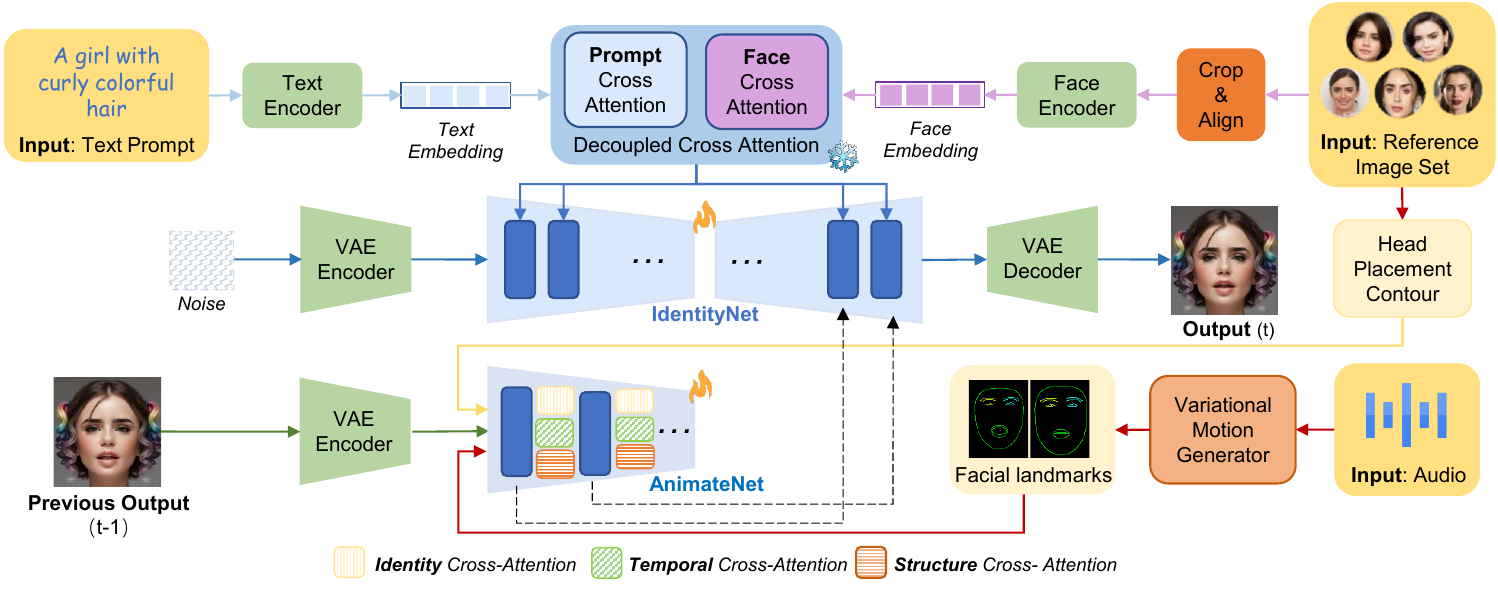}
    \caption{PortraitTalk has two main components: IdentityNet and AnimateNet. Text and identity embeddings are derived from the text and face encoder, with a projection layer mapping identity features into the text embedding dimension. These features are integrated into IdentityNet using a decoupled cross-attention mechanism to capture subtle facial characteristics. Simultaneously, facial motions corresponding to the input speech, enhanced by head placement guidance, are processed through AnimateNet to ensure dynamic and temporal coherence. In PortraitTalk, a latent diffusion model serves as the foundational rendering mechanism. The structural attention block incorporates head placement guidance and facial landmark mapping. For simplicity, these elements are represented within a single block.
    }
    \label{fig1}
\end{figure*}

Recently, Diffusion Models (DM~\cite{105596298DM}) have been increasingly adopted for talking face generation~\cite{stypulkowski2024diffused,wei2024aniportrait, wang2024vexpressconditionaldropoutprogressive,xu2024hallohierarchicalaudiodrivenvisual, chen2024echomimiclifelikeaudiodrivenportrait}, offering improvements in both visual quality and audio-lip synchronization. Despite these advances, such models often struggle with generalization, particularly when presented with identities that differ significantly from those seen during training. Some approaches~\cite{wei2024aniportrait,wang2024vexpressconditionaldropoutprogressive}, exhibit limited audio-lip synchronization, which undermines the naturalness of the output. Furthermore, fine-grained control and customization in audio-driven talking face generation remain underexplored in existing diffusion-based models.

\subsection{Emotional Talking Face Generation}
Emotional talking face generation has attracted increasing attention, primarily due to its potential to enhance communication and its utility in the entertainment industry. Traditional methods have predominantly used discrete emotion labels to model expressions. 
For example,~\cite{EmoGen} developed a model inspired by Wav2Lip~\cite{wav2lip}, integrating an emotion label encoder, emotion-specific loss function, and an emotion discriminator. This model, however, primarily modifies the mouth region, thereby producing videos that lack realistic appearance variations. Furthermore, it exhibits limited effectiveness in displaying diverse expressions and fails to maintain an identity consistency across the generated frames.

%Sun_Xuan_Liu_Xiang_2024, 
Label-based emotional talking face generation typically struggles to produce controllable and fine expressions. To address these limitations, recent developments advocate the use of emotional videos as references~\cite{Sun_Xuan_Liu_Xiang_2024,tan2024say,EAMM,edtalk,10816597}. This approach allows for the injection of expressions directly derived from other videos, enhancing the expressiveness of generated faces. For example, EAMM~\cite{EAMM} uses an emotional source video to generate emotional talking faces. Nonetheless, this method often struggles with identity consistency, exhibits noticeable irregularities, and encounters audio-lip synchronization issues. 
These challenges underscore the necessity for the development of a model that can integrate emotion with audio-visual semantics more effectively. 
Similarly, EDTalk~\cite{edtalk} employs emotions from reference-style videos to enhance the expressiveness of talking faces. While this method substantially enhances expressiveness, it sometimes does so at the cost of visual quality.

\section{The Proposed Method}
\label{sec_method}
PortraitTalk is a customizable audio-driven talking face generation framework that synthesizes identity-preserving facial animations with accurate audio-lip synchronization, using reference images and the corresponding speech. Fig.~\ref{fig1} illustrates the pipeline of PortraitTalk. To enhance customization and controllability, we incorporate text prompt embeddings, enabling the model to edit facial attributes such as style, expression, and appearance during the generation process. 
The framework consists of two main components. The first, IdentityNet, extracts both high-level semantic information and low-level visual details from the reference image and text prompt, ensuring consistent identity preservation across frames. The second, AnimateNet, conditions the generation on speech-driven motion features and employs temporal mechanisms to maintain frame-to-frame coherence and enable smooth, realistic transitions throughout the video.

\subsection{IdentityNet}
\label{identitynet}
IdentityNet is built upon the architecture of Stable Diffusion~\cite{SDiffusion} (SD 1.5) and serves as the primary component for preserving facial identity throughout the generated video. The backbone is initialized with the pre-trained UNet weights from SD 1.5, enabling the model to retain the strong generative capacity of the original network while adapting to the talking face generation task through fine-tuning on our dataset. This initialization provides a stable foundation for IdentityNet. 

\begin{figure}[!t]
    \centering
    \includegraphics[trim={4cm 22cm 3cm 2cm}, clip, width=1\linewidth]{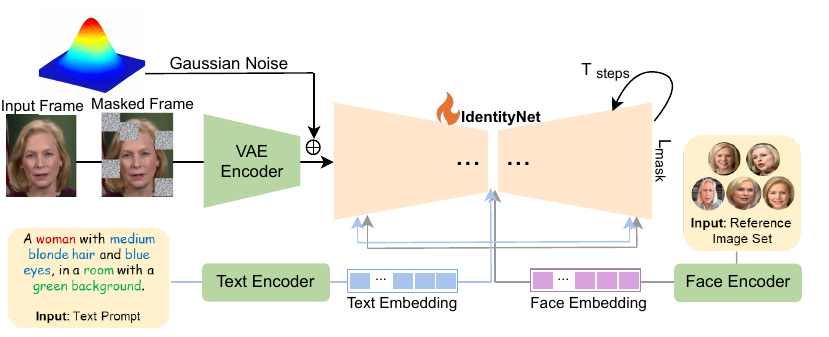}
    \caption{
    An overview of the masked loss fine-tuning strategy used for IdentityNet. During training, random regions of the input frames are corrupted, and the model is optimized to reconstruct the original content. This masked fine-tuning approach encourages the network to focus on the global facial structure and identity-relevant features, rather than overfitting to local pixel-level details. The process operates in the latent space of a diffusion model: the masked input is first encoded, followed by a forward and backward diffusion process over multiple time steps ($T_{steps}$). The denoising U-Net is trained to estimate the added noise using a mask reconstruction loss ($L_{mask}$), which guides IdentityNet toward producing more consistent, stable, and identity-preserving generations.
    }
    \label{fig1_2}
\end{figure}

To improve the consistency of synthesized frames and enhance the reconstruction capability of IdentityNet, we apply a masked fine-tuning strategy, as depicted in Fig.~\ref{fig1_2}. Specifically, random regions of the input frames are corrupted with Gaussian noise, and the model learns to reconstruct the missing content. Unlike conventional zero-masking approaches, the use of Gaussian noise aligns more naturally with the denoising objective of diffusion models. This strategy encourages the model to attend to global facial structure, such as head shape and overall appearance, rather than overfitting to local pixel patterns, thereby promoting more stable and identity-consistent generations~\cite{Chen_2024_CVPR}.

The fine-tuning process follows the training strategy of stable diffusion. Randomly masked frames are encoded by passing through the encoder to obtain a masked latent representation $\mathbf{z}_{m}$. Subsequently, a forward and backward diffusion process is applied across \(T\) time steps within this masked latent space. The denoising U-Net ($\epsilon_{\theta}$) is trained to reconstruct the original image by estimating the amount of noise ($\epsilon$) introduced at each forward step. This is achieved by the mask reconstruction loss:
\begin{equation}
    L^{mask} = \mathbb{E}_{t,\mathbf{z}_{t}, c, \epsilon \sim \mathcal{N}(0,1)}\left[ \| \epsilon - \epsilon_\theta(\mathbf{z}_{m_{t}}, t, \mathbf{c}) \|^2 \right]
\end{equation}
where $t$ is a sampled time step, ${z}_{m_{t}}$ is the noisy masked latent at time step $t$, and $c$ is the condition set (text and reference images). 
This loss function guides the network towards accurate noise estimation and effective image restoration.  

Despite the effectiveness of SD in image generation, it struggles to maintain consistent identity characteristics across consecutive frames. Moreover, given our model's goal of enabling customization during the generation process, it is essential to incorporate user prompts, both textual and visual, into the denoising blocks effectively. This ensures that the generated talking faces consistently exhibit facial details that align with the input text and reference image. 
To tackle these challenges, we employ a decoupled cross-attention mechanism that processes text and image features independently, enabling more targeted and coherent feature integration. 
Specifically, we apply the decoupled cross-attention module~\cite{ye2023ipAdapter} to process text and reference image features independently, embedding semantic representations from both modalities into the IdentityNet backbone to guide the generation process.

In preceding studies~\cite{ye2023ipAdapter,li2023photomaker}, CLIP has been widely used as an image encoder. However, it has proven inadequate for extracting strong facial features in audio-to-talking face generation. 
This deficiency often results in unstable facial representations across frames, while barely reflecting text prompt guidance in generated videos. CLIP, a contrastive language–image model, is primarily trained on text-image pairs from the natural image domain. This training predisposes CLIP to prioritize broad and ambiguous features such as color and style, which are less effective for detailed face image understanding~\cite{wang2024instantid}. Such limitations pose significant challenges for the proposed framework. 
This task necessitates precise identity preservation, wherein robust semantic representation of facial features is crucial. 
To achieve this, we initially eliminate background elements from reference images and subsequently extract facial embeddings utilizing a face recognition model~\cite{wang2024instantid}. This approach has demonstrated enhanced facial fidelity across generated frames. 
Simultaneously, text prompts are processed using the CLIP text encoder, and facial embeddings are mapped to the same dimensional space as text embeddings via a projection layer. Both text and facial representations are then integrated into the denoising UNet using decoupled cross-attention blocks, enhancing the facial fidelity and editability of the generated talking faces. 

As IdentityNet is primarily designed to preserve the facial details of a specific identity, it does not inherently maintain the temporal consistency essential for video generation. Therefore, we introduce AnimateNet to mitigate this issue.

\subsection{AnimateNet}
\label{animateNet} 
Enforcing accurate facial motion is essential for generating realistic talking face videos. To address this, we divide the animation process into two phases: audio-to-motion and motion-to-frame generation. In the audio-to-motion phase, a sequence of facial movements (represented as facial landmarks) is generated to capture the expected lip and facial dynamics. This is achieved using an audio-to-motion module~\cite{ye2023geneface++}, which combines a variational autoencoder with the HuBERT audio Transformer~\cite{HuBERT}. The module employs a dilated convolutional encoder and decoder to improve the robustness of feature extraction and support the generation of longer, coherent motion sequences. Following generation, the facial motion representations are refined to align with MediaPipe~\cite{mediapipe} landmarks, further enhancing AnimateNet’s training stability and generalization across diverse speaking styles.

The motion-to-frame generation phase is responsible for producing realistic talking faces aligned with the motion features extracted from the audio. 
To achieve this, we introduce AnimateNet, built upon SD 1.5 and inspired by the design principles of ControlNet~\cite{controlnet}. In ControlNet, an additional trainable branch is introduced alongside a frozen diffusion model to condition the generation on external inputs without disrupting the original generative capabilities. Similarly, AnimateNet integrates trainable control branches that modulate the generation process based on motion-related conditions, enabling the model to faithfully synthesize speech-driven facial movements. Motion features extracted by AnimateNet are fused with IdentityNet through ZeroConv layers, allowing seamless integration of motion and appearance information without disrupting the learned representations of either network. 
Additionally, AnimateNet employs three distinct cross-attention mechanisms (Structure, Identity, and Temporal), each designed to process a specific condition.

The \textbf{structure cross-attention}, processes facial landmarks corresponding to the associated audio segment, ensuring accurate alignment of audio-lip movements. Next, \textbf{identity cross-attention} focuses on head placement guidance, which is derived from the extracted head contour of the previous talking head or, for the initial frame, the reference image. To obtain this guidance, we extract the head contour from each frame using a modified Canny edge detector with reduced sensitivity to local details. This produces a coarse binary edge map that captures the overall shape and position of the head while ignoring fine-grained facial features. We encode the contour maps through a shallow convolutional branch and inject them into the diffusion via ZeroConv layers. This allows the network to condition the generation on head placement without interfering with identity or descriptive features. 
This approach addresses the challenge of variability in head movements, attributable to the stochastic nature of diffusion models, ensuring that head movements remain consistent and realistic across frames. 
To ensure smooth transitions between frames, and inspired by prior work~\cite{tian2024emoemoteportraitalive,stypulkowski2024diffused}, we condition the generation at each time step on the frame produced in the previous step using a \textbf{temporal cross-attention} mechanism. 
Finally, a weighted sum of each attention block ($Z^{\text{new}}$) is computed and serves as the input for subsequent layers. The attention computation can be indicated as: 
\begin{align}
 \label{eq2}
   Z^{\text{new}} = & \, w_{1} \left( \text{CrossAttn}_{\textbf{Structure}}(Q,K_{1},V_{1}) \right) \nonumber \\
   & + w_{2} \left( \text{CrossAttn}_{\textbf{Identity}}(Q,K_{2},V_{2}) \right) \nonumber \\
   & + w_{3} \left( \text{CrossAttn}_{\textbf{Temporal}}(Q,K_{3},V_{3}) \right)
\end{align}
where $\text{CrossAttn}(Q,K_{i},V_{i})$ computes the attention score following the standard attention~\cite{NIPS2017_3f5ee243} mechanism. $Q$, $K$ and $V$ represent the query, key and value matrices of the attention process. While the query remains shared across conditions, each condition-specific branch learns independent key-value mappings. The relative importance of each condition is controlled via learnable weights ($w_i$). 
The outputs of these attention mechanisms are integrated into IdentityNet, where the weighted sum of the resulting feature maps is used for the final generation step. 

For training, AnimateNet optimizes a denoising diffusion objective similar to stable diffusion:
\begin{equation}
    L = \mathbb{E}_{t,\mathbf{z}_{t}, c, \epsilon \sim \mathcal{N}(0,1)}\left[ \| \epsilon - \epsilon_\theta(\mathbf{z}_{t}, t, \mathbf{c}) \|^2 \right]
\end{equation}
where $c$ refers to the set of conditions, including facial landmarks, head placement guidance and temporal information. By decoupling condition learning and leveraging specialized cross-attention mechanisms, AnimateNet achieves robust temporal consistency, natural head movements, and accurate audio-lip synchronization, resulting in realistic and identity-consistent talking face videos.

\subsection{Audio-Driven Facial Dynamics Score}
Existing metrics in audio-to-talking face generation often focus on individual aspects of talking face video (\textit{e.g.,} visual quality~\cite{10378582}, audio-lip synchronization~\cite{Chung16a}). 
There is a need for a comprehensive metric that evaluates realistic facial details, temporal alignment, and motion coherence. 
To this end, we advocate Audio-Driven Facial Dynamics (ADFD) that assesses both the spatial alignment and dynamic movement of facial landmarks over time, advancing beyond traditional metrics by incorporating these additional factors relative to the audio. 
The ADFD metric is formulated as:
\begin{align}
     \quad ADFD &  = w_1 \left( \frac{1}{T} \sum_{t=1}^T \left(1 - \sqrt{\sum_{i=1}^n (L_i^{gen} - L_i^{gt})^2} / d \right) \right)
     \nonumber \\
    &\quad \times w_2 \left( \frac{ \left(\mathbf{M}_t^{gen} \cdot \mathbf{M}_t^{gt} \right) + 1}{2\|\mathbf{M}_t^{gen}\| \|\mathbf{M}_t^{gt}\|} \right)
\end{align}
where, $T$ represents the total number of frames in a video, $L_{i}^{gen}$ and $L_{i}^{gt}$ are the arrays of generated and ground truth landmarks, respectively, at each time step $t$. $d$ is the maximum distance between two points in the frame, which is used for normalization. Motion vectors of the generated and ground truth landmarks at time step $t$ are represented by $M_{t}^{gen}$ and $M_{t}^{gt}$. To determine these motion vectors, we calculate the difference between two consecutive frames. $w_1$ and $w_2$ are the balancing parameters. 

The ADFD score has two main parts. The first quantifies the spatial alignment of facial landmarks between generated and ground-truth talking face videos by measuring the normalized Euclidean distance between the corresponding landmarks. The second part assesses the motion coherence and temporal consistency of the facial movements by calculating the cosine similarity of the motion vectors. This part evaluates the extent to which the direction and scale of facial movements in the generated video correspond with those in the ground truth over time. 
To accommodate specific research needs, ADFD can be fine-tuned using adjustable weights ($w_1$ and $w_2$), allowing one to prioritize certain aspects of the evaluation. 
Further, to ensure that all values reflect degrees of similarity on a non-negative scale, the range of the cosine similarity component has been adjusted from $[-1,1]$ to $[0,1]$. 
A high value in the ADFD score (close to $1$) signifies that the generated talking face video not only aligns well spatially but also demonstrates temporal consistency across frames.

\section{Experimental Results}
\label{sec_exp}
\subsection{Datasets and Evaluation Metrics}
We use the HDTF~\cite{zhang2021flow} and MEAD\cite{kaisiyuan2020mead} datasets for evaluation. 
The HDTF dataset contains high-quality videos of over 300 identities from YouTube.
For training, we randomly select 6 videos, approximately $30,000$ frames in total. 
MEAD contains $60$ speakers with 8 different emotions. This dataset is captured in a laboratory setting and each emotion is captured under three different emotion intensity levels. 
In this paper, four randomly selected videos from MEAD are used for training. 
For the testing stage, similar to~\cite{edtalk}, we randomly select $20\%$ of the HDTF dataset and choose speakers `M003', `M009', `M030', and `W015' from MEAD. Since both HDTF and MEAD lack text prompts, we manually created descriptive prompts for the selected videos used in fine-tuning, capturing key attributes such as gender, background, and hair color. 

\begin{table*}[!t]
\centering
\label{table1}
\resizebox{\textwidth}{!}{%
\begin{tabular}{lccccc:c|ccccc:cc}
\toprule
\textbf{Method} & \multicolumn{6}{c|}{\textbf{MEAD}~\cite{kaisiyuan2020mead}} & \multicolumn{6}{c}{\textbf{HDTF}~\cite{zhang2021flow}} \\
\cmidrule(lr){2-7} \cmidrule(lr){8-13}
& \textbf{PSNR↑} & \textbf{SSIM↑} & \textbf{M/F-LMD↓} & \textbf{FID↓} & \textbf{SyncNet ↑} & \textbf{ADFD ↑} & \textbf{PSNR↑} & \textbf{SSIM↑} & \textbf{M/F-LMD↓} & \textbf{FID↓} & \textbf{SyncNet ↑} & \textbf{ADFD ↑} \\
\midrule
MakeItTalk~\cite{makeittalk} & 19.442 & 0.614 & 2.541/2.309 & 37.917 & 5.176 & 0.693 & 21.985 & 0.709 & 2.395/2.182 & 18.730 & 4.753 & 0.726 \\
Wav2Lip~\cite{wav2lip}    & 19.875 & 0.633 & 1.438/2.138 & 44.510 & 8.774 & 0.584 & 22.323 & 0.727 & 1.759/2.002 & 22.397 & \textbf{9.032} & 0.699 \\
Audio2Head~\cite{wang2021audio2head} & 18.764 & 0.586 & 2.053/2.293 & 27.236 & 6.494 & 0.725 & 21.608 & 0.702 & 1.983/2.060 & 29.385 & 7.076 & 0.835 \\
SadTalker~\cite{Zhang_2023_CVPR}  & 19.042 & 0.606 & 2.038/2.335 & 39.308 & 7.065 & 0.761 & 21.701 & 0.702 & 1.995/2.147 & 14.261 & 7.414 & 0.726 \\
IP-LAP~\cite{Zhong_2023_CVPR}     & 19.832 & 0.627 & 2.140/2.116 & 46.502 & 4.156 & 0.527 & 22.615 & 0.731 & 1.951/1.938 & 19.281 & 3.456 & 0.633 \\
TalkLip~\cite{wang2023seeing}    & 19.492 & 0.623 & 1.951/2.204 & 41.066 & 5.724 & 0.548 & 22.241 & 0.730 & 1.976/1.937 & 23.850 & 1.076 & 0.518 \\
EAMM~\cite{EAMM}  & 18.867 & 0.610 & 2.543/2.413 & 31.268 & 1.762 & \textbf{0.847} & 19.866 & 0.626 & 2.910/2.937 & 41.200 & 4.445 & \textbf{0.839} \\
EDTalk~\cite{edtalk}    & 21.628 & 0.722 & 1.537/\textbf{1.290} & 17.698 & 8.115 & 0.715 & 25.156 & 0.811 & 1.676/1.315 & 13.785 & 7.642 & 0.722 \\
\textbf{PortraitTalk (Our Model)} & \textbf{23.097} & \textbf{0.873} & \textbf{1.206}/1.385 & \textbf{17.351} & \textbf{8.916} & 0.816 & \textbf{27.495} & \textbf{0.846} & \textbf{1.157}/\textbf{1.017} & \textbf{11.753} & 8.381 & 0.835\\
\bottomrule
\end{tabular}
}
\vspace{0.1cm}
\caption{A quantitative comparison with the SOTA methods on MEAD and HDTF. The highest scores for each metric are highlighted in bold. The symbols '↑' and '↓' denote that higher and lower values, respectively, indicate better performance.}
\label{table1}
\end{table*}

Following previous work~\cite{xu2024hallohierarchicalaudiodrivenvisual,chen2024echomimiclifelikeaudiodrivenportrait}, we adopt established metrics such as PSNR~\cite{PSNR}, SSIM~\cite{SSIM}, FID~\cite{NIPS2017_8a1d6947}, E-FID~\cite{chen2024echomimiclifelikeaudiodrivenportrait}, and FVD~\cite{unterthiner2019accurategenerativemodelsvideo} to evaluate the visual quality and generative performance of our model. To assess facial motion accuracy and temporal consistency, we include Landmark Distance (LMD)~\cite{LMD} and the proposed Audio-Driven Facial Dynamics (ADFD) score. The SyncNet error metric is used to measure audio-visual synchronization, which is essential for evaluating lip-sync accuracy in talking face generation. In addition, we use CLIP-T~\cite{pmlrv139radford21a}, DINO~\cite{Caron_2021_ICCV}, and Face Similarity~\cite{li2023photomaker} to assess perceptual relevance and identity preservation. CLIP-T evaluates consistency with the input prompt, DINO captures visual structure, and Face Similarity measures how closely the generated face matches the reference identity.

\subsection{Implementation Details}
Our framework, built upon Stable Diffusion v1.5~\cite{Rombach2021HighResolutionIS}, leverages the $RealisticVision$ weights to fine-tune the model for audio-to-talking face generation. The implementation is carried out in PyTorch.
We train our model using the AdamW~\cite{Loshchilov2017DecoupledWD} optimizer with a batch size of $32$ and a learning rate of $5e-6$. 
The training process encompasses a multi-stage approach. 
Initially, each component is trained independently. IdentityNet is finetuned for 10 epochs, followed by the training of AnimateNet (excluding temporal attention) for $100$ epochs. Subsequently, in the second stage, the entire framework undergoes integrated training for $50$ epochs. All the training and testing phases are conducted on a single NVIDIA A100 GPU. 
The overall training takes approximately two weeks.

\subsection{Quantitative Results}
\label{Quantitative_Results}
To assess our model, we conducted a quantitative comparison with the state-of-the-art methods in audio-to-talking-face generation. The results are reported in Table~\ref{table1}. PortraitTalk consistently outperforms the existing approaches across nearly all the metrics on the HDTF and MEAD datasets. Wav2Lip~\cite{wav2lip} achieves a relatively higher SyncNet score on HDTF, primarily because it utilizes this metric as one of its training loss functions. 
Our model achieves a satisfactory SyncNet score, surpassing Wav2Lip on the MEAD dataset, and excels in audio-lip alignment and facial structure accuracy, as indicated by LMD and ADFD metrics. 
Although EAMM has the highest ADFD score, our model ranks closely behind on the HDTF dataset, showing similar effectiveness. The slightly larger gap on the MEAD dataset is likely due to its wider range of emotional expressions and varying intensities, which directly impact lip movements. 

\begin{figure}[!t]
    \centering
    \includegraphics[trim={0cm 0.7cm 17cm 0cm}, clip, width=1\columnwidth]{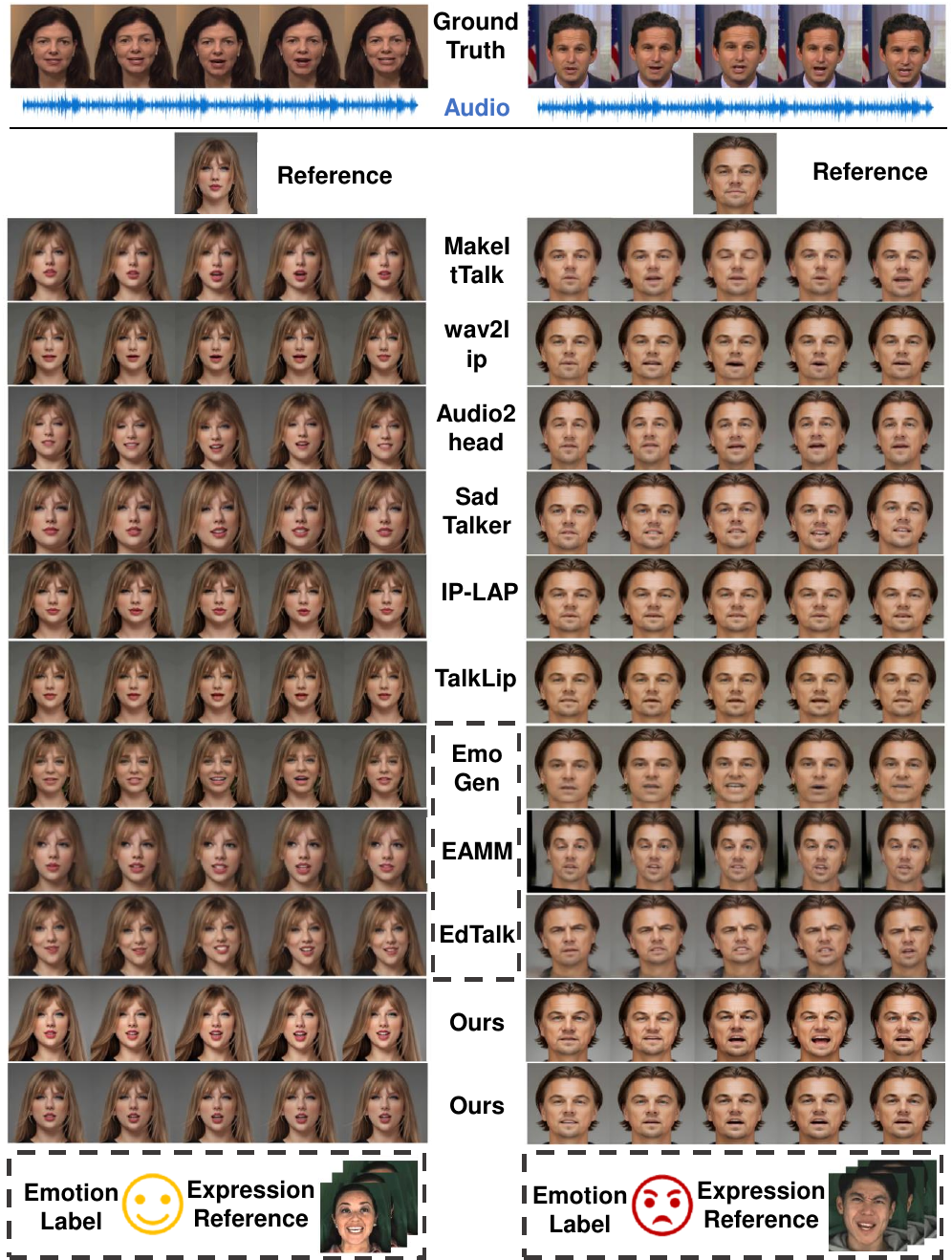}
    \caption{Qualitative comparison with the existing talking face generation methods. 
    The results demonstrate that PortraitTalk surpasses previous methods in audio-lip alignment, identity resemblance, and expressiveness. 
    Please note that the methods in the dash box use external emotion labels or reference videos to generate expressive videos.}
    \label{fig2}
\end{figure}

\begin{figure}[!t]
    \centering
    \includegraphics[trim={0cm 0.1cm 0cm 0cm}, clip, width=1\columnwidth]{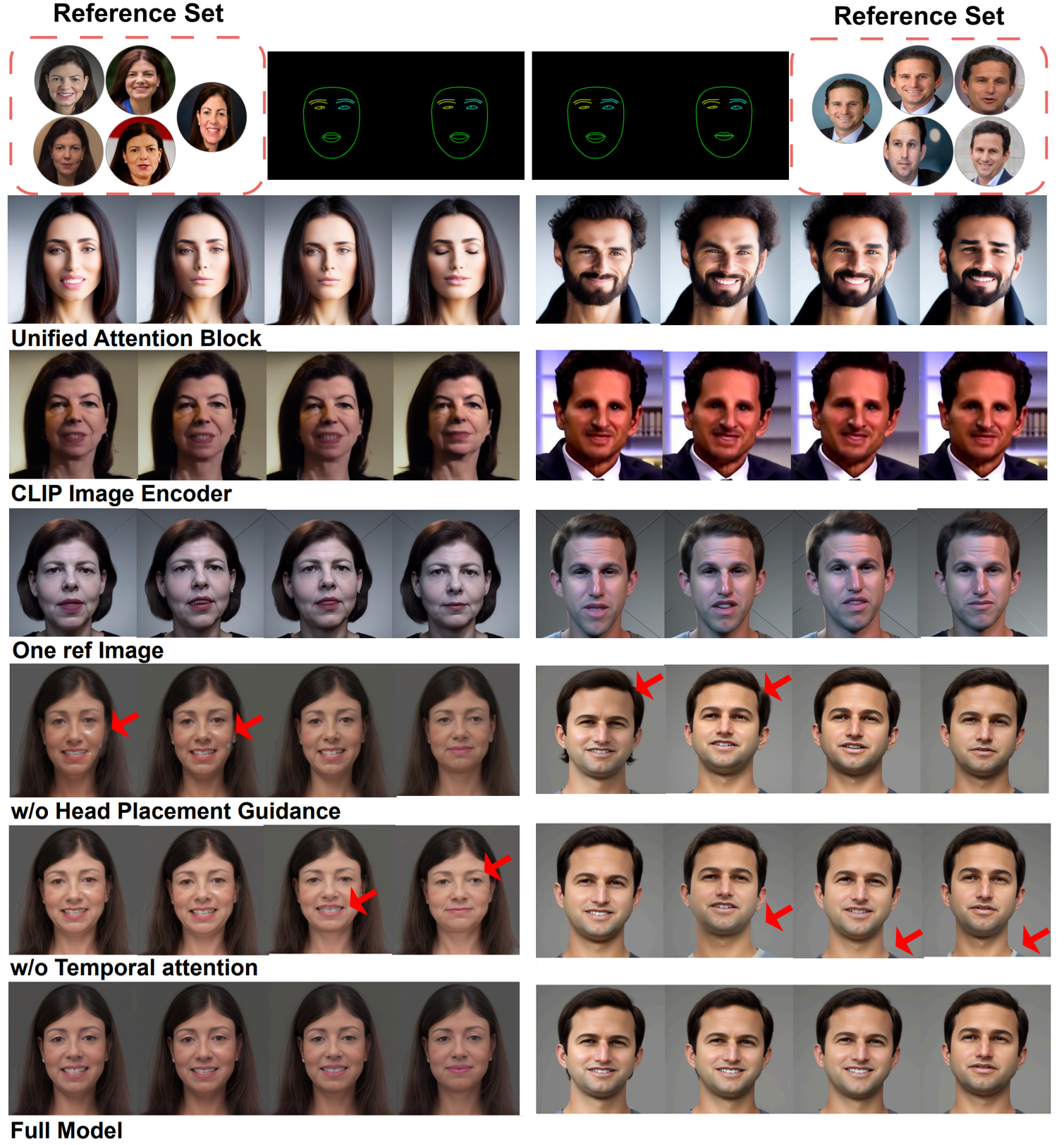}
    \caption{Qualitative comparison of ablated variants of PortraitTalk, illustrating the effect of different model components. Each column shows output frames generated by models with specific components removed or added. These visualizations demonstrate the contribution of each design choice to the final visual quality and identity consistency of the generated talking head videos. This ablation study highlights how each component contributes to enhancing realism, expressiveness, and alignment with user intent.
    }
    \label{fig3}
\end{figure}

\subsection{Qualitative Results}
In Fig.~\ref{fig2}, we compare the generation quality of PortraitTalk against the state-of-the-art approaches. 
MakeItTalk and Wav2Lip produce low-quality outputs with artifacts, notably around the mouth, which fail to blend seamlessly with adjacent facial features.
Audio2Head~\cite{wang2021audio2head}, TalkLip~\cite{wang2023seeing}, and IP-LAP~\cite{Zhong_2023_CVPR} exhibit shortcomings in achieving precise audio-lip synchronization, with generated mouth movements often slightly open and displaying distortions across frames, thereby diminishing the realism of the output.

In the emotional talking face generation, existing methods typically rely on discrete emotional labels or facial expressions from user-provided reference videos, which come with notable limitations. For instance, EAMM~\cite{EAMM} and EmoGen~\cite{EmoGen} struggle with identity consistency, as shown in Fig.~\ref{fig2}, where EAMM fails for the male subject and EmoGen for the female subject. Additionally, EmoGen restricts modifications to a facial square region, resulting in blurred areas that compromise realism. While EDTalk~\cite{edtalk} improves emotional generation, it still suffers from jitter and irregular mouth shapes in frames. 
In contrast, our model demonstrates superior performance in both emotion-agnostic and emotional talking face generation. It excels at preserving identity consistency, ensuring precise audio-lip synchronization, and maintaining consistent head movements throughout the video.

\begin{table*}[!t]
    \centering

        \begin{tabular}{lccccc}
            \toprule
            \textbf{Metric/Method} & \textbf{PSNR↑} & \textbf{SSIM↑} & \textbf{FID↓} & \textbf{SyncNet↑}& \textbf{ADFD↑} \\ \midrule
            Unified attention block & 9.430 & 0.278 & 12.794 & 0.0412 & 0.024 \\
            w/o Face Encoder & 13.457 & 0.490 & 15.794 & 2.912 & 0.386 \\
            One reference image & 16.296 & 0.593 & 14.551 & 2.714 & 0.582 \\
            w/o head placement guidance & 24.725 & 0.748 & 13.821 & 6.264 & 0.683 \\ 
            w/o temporal attention & 26.951 & 0.796 & 12.381 & 7.486 & 0.765\\ 
            \textbf{Full model} & 27.495 & 0.846 & 11.753 & 8.381 & 0.835\\ 
            \bottomrule
        \end{tabular}
    \vspace{0.1cm}
    \caption{Quantitative results of the ablation study evaluating the impact of different components in PortraitTalk. The results show that excluding key elements such as temporal attention, head placement guidance, and the face encoder leads to notable drops in performance, particularly in synchronization and identity preservation metrics. The full model consistently achieves the best performance across all metrics, confirming the effectiveness of our complete system design.
    }
    \label{table2}
\end{table*}

\subsection{Ablation Study}
\label{ablationStudy}
To evaluate the critical components of our framework, we conduct an ablation study in this part. 

\textbf{Unified Attention Block:}
Initially, we merge the separate cross-attention blocks of IdentityNet into a single attention block to assess its impact on performance. The results are shown in Fig.~\ref{fig3} and Table~\ref{table2}. The findings indicate a significant degradation in the model's ability to preserve the intended identity, with its recognition capability reduced to identifying only the gender from text prompts. 

\textbf{Impact of the Face Encoder:}
Further, we substitute our dedicated face encoder with the CLIP image encoder to explore the influence of the encoding mechanism on the generation capability of PortraitTalk. 
While there is an observable enhancement in image fidelity and identity consistency compared to the initial ablation scenario, the model still struggles to maintain detailed identity attributes across frames accurately. 
These experiments validate our decision to employ a specific face encoder optimized for facial feature extraction, underscoring its importance in achieving high-quality and consistent talking face generation. 

\begin{table}[!t]
    \centering
        \begin{tabular}{lcccc}
            \toprule
            \textbf{Metric/Method} & \textbf{FID↓} & \textbf{FVD↓} & \textbf{SSIM↑} & \textbf{E-FID↓}\\ \midrule
            Aniportrait~\cite{wei2024aniportrait} & 53.72 & 1042.93 & 0.731 & 1.94 \\
            V-Express~\cite{wang2024vexpressconditionaldropoutprogressive} & 59.34 & 1173.25 & 0.716 & 1.81\\
            Hallo~\cite{xu2024hallohierarchicalaudiodrivenvisual} & 37.81 & 583.07 & 0.768 & 1.73  \\
            Echomimic~\cite{chen2024echomimiclifelikeaudiodrivenportrait} & 32.51 & \textbf{536.41} & 0.775 & 1.68 \\ 
            \textbf{PortraitTalk} & \textbf{31.86} & 586.19 & \textbf{0.786} & \textbf{1.64} \\ 
            \bottomrule
        \end{tabular}
    \vspace{0.1cm}
    \caption{Performance comparison of audio-driven talking face generation models using standard visual and temporal metrics.}
    \label{table3}
\end{table}

\textbf{Number of Reference Images:}
We investigate the impact of using multiple reference images on the quality and realism of generated visual content. As demonstrated in Fig.~\ref{fig3}, employing multiple images of the same identity, showcasing various expressions or head angles, significantly enhances the model's capability to learn the underlying facial structure of a person. This strategy is essential for facilitating customization in generated talking face videos, enabling the model to respond to a variety of customization prompts effectively. In contrast, reliance on a single reference image leads to a considerable reduction in both the quality and realism of the generated faces. 
This issue is especially evident in the male case (Fig.~\ref{fig3}), where the model produces an overly oval head shape.

\textbf{Impact of Structure and Temporal Attention:}
We also try to remove the head placement guidance and temporal attention mechanisms from AnimateNet. This modification leads to a noticeable decrease in coherence across the generated video frames. As illustrated in Fig.~\ref{fig3}, while the lip synchronization remains somewhat accurate, there is a clear inconsistency in head placement and noticeable distortions between consecutive frames. Interestingly, without temporal attention, the generative model often fails to maintain consistency in the clothing of the generated talking faces. This variation underscores the crucial role of head placement guidance and temporal attention in maintaining the stability and continuity of head movements throughout the video.

\begin{figure}[!t]
    \centering
    \includegraphics[trim={0.7cm 11.5cm 8cm 1cm}, clip, width=.9\columnwidth]{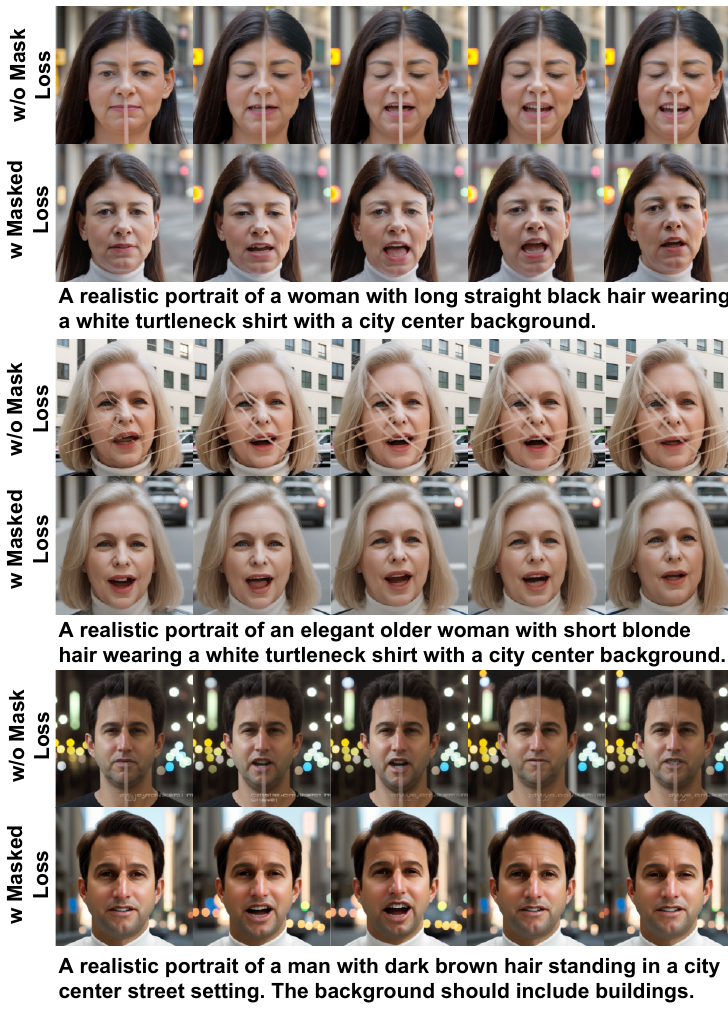} 
    \caption{Qualitative impact of mask loss on talking face generation. When mask loss is not applied during training, the model is more prone to generating artifacts such as text-like distortions and structural inconsistencies. Incorporating mask loss encourages the network to learn visual consistency by recovering corrupted regions, resulting in cleaner, more stable, and visually coherent frames.
    }
    \label{fig8}
\end{figure}

\subsection{Comparison with Recent Diffusion-based models}
We compare PortraitTalk against recent diffusion-based audio-driven talking face generation frameworks, including AniPortrait~\cite{wei2024aniportrait}, V-Express~\cite{wang2024vexpressconditionaldropoutprogressive}, Hallo~\cite{xu2024hallohierarchicalaudiodrivenvisual}, and EchoMimic~\cite{chen2024echomimiclifelikeaudiodrivenportrait}. As shown in Table~\ref{table3}, while EchoMimic achieves the lowest FVD score, indicating stronger temporal consistency, PortraitTalk demonstrates a more balanced performance across all evaluation metrics. In particular, PortraitTalk obtained superior results in visual fidelity (FID, SSIM) and expression fidelity (E-FID), while maintaining competitive temporal coherence, leading to consistent identity preservation and realistic motion. 
It is also worth noting that PortraitTalk provides the flexibility to customize reference identity attributes, style, and the overall scene during generation, a capability not supported by current state-of-the-art methods. Visual examples of this functionality are presented in Sections~\ref{prompt_conflict} and~\ref{application}.

\begin{table*}[!t]
    \centering
        \begin{tabular}{lcccc}
            \toprule
            \textbf{Method} & \textbf{CLIP-T\% ↑} & \textbf{DINO\% ↑} & \textbf{Face.sim\% ↑} & \textbf{SyncNet score↑}\\ 
            \midrule
            w/o mask loss & 20.53 & 57.8 & 56.2 & 6.51 \\ 
            \textbf{w mask loss} & 20.54 & 64.1 & 63.5 & 8.36 \\ 
            \bottomrule
        \end{tabular}
    \vspace{0.1cm}
    \caption{Evaluation of the impact of mask loss on talking face generation. 
    Results are reported across perceptual alignment (CLIP-T), structural consistency (DINO), identity preservation (Face Similarity), and audio-visual synchronization (SyncNet).
    }
    \label{table4}
\end{table*}

\begin{figure}[!t]
    \centering
    \includegraphics[trim={0.0cm 13.3cm 5cm 1cm}, clip, width=\columnwidth]{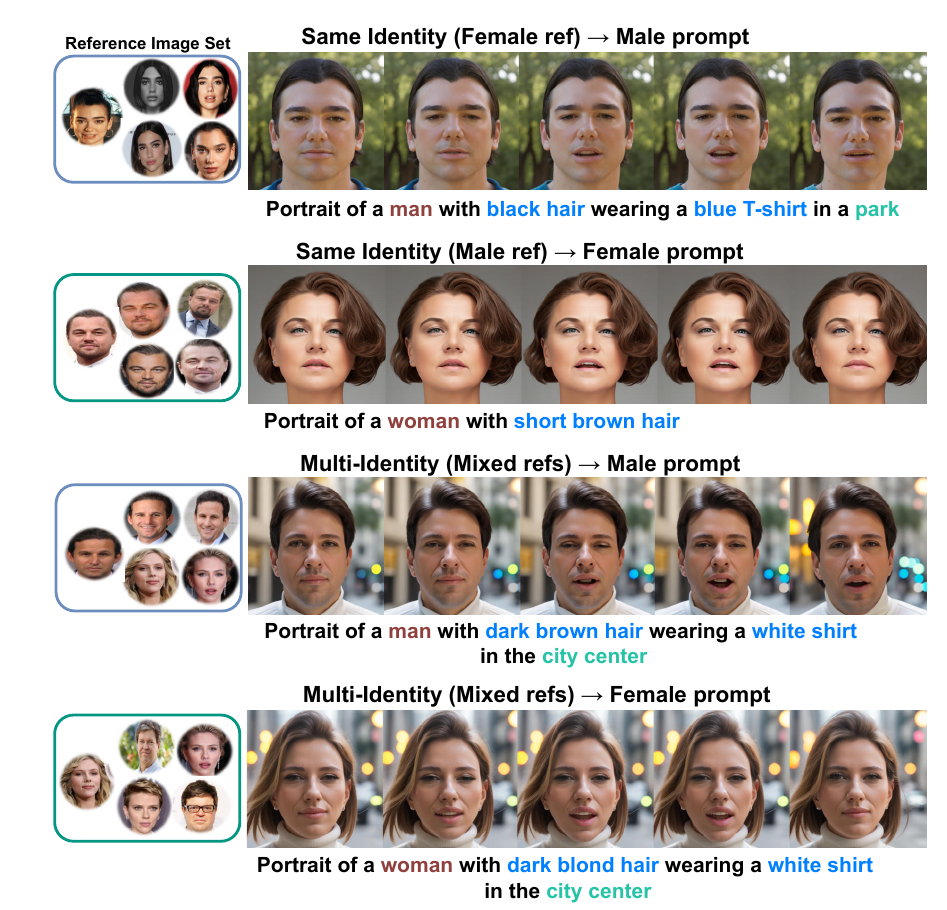} 
    \caption{Balancing the contributions of reference images and text prompts in PortraitTalk. Our model demonstrates robustness in handling conflicting inputs, such as when the visual identity in the reference image does not align with the subject described in the prompt. In these cases, PortraitTalk adapts the generated face to reflect the textual description while preserving structural coherence from reference images. Additionally, by using multiple references, including from different identities, the model can blend features to synthesize coherent and novel outputs.
    }
    \label{fig7}
\end{figure}

\subsection{Impact of The Mask Loss}
In this section, we evaluate the impact of incorporating mask loss during the training of PortraitTalk. As shown in Table~\ref{table4}, applying mask loss improves the overall generation quality, particularly in terms of identity preservation, visual fidelity, and audio-lip synchronization.  By learning to recover corrupted regions, the model implicitly strengthens its understanding of visual consistency, leading to reduced artifacts and more stable frame-wise outputs.
As illustrated in Fig.~\ref{fig8}, this approach also mitigates unintended artifacts such as text-like distortions or detail inconsistencies issues often caused by the inherited biases of the underlying diffusion model. 

\subsection{Prompt and Reference Conflicts}
\label{prompt_conflict}
PortraitTalk provides flexible control over both visual identity and text-driven attributes during generation. 
As illustrated in Fig.~\ref{fig7}, PortraitTalk remains robust even when the reference image and prompt convey conflicting information. 
For example, when provided with a female reference image and a prompt describing a male subject, the model adapts facial features accordingly, generating outputs that reflect the textual description while maintaining structural consistency from the reference images. 
Furthermore, the multi-reference design offers additional flexibility. By supplying reference images from multiple identities, including across genders, the model can blend facial features to synthesize new, coherent identities. This demonstrates PortraitTalk’s ability to generalize beyond one-to-one mappings and to generate diverse, controllable outputs based on cross-modal and multi-source inputs.

\begin{figure}[!t]
    \centering
    \includegraphics[trim={0cm 1.2cm 17.5cm 0cm}, clip, width=.95\columnwidth]{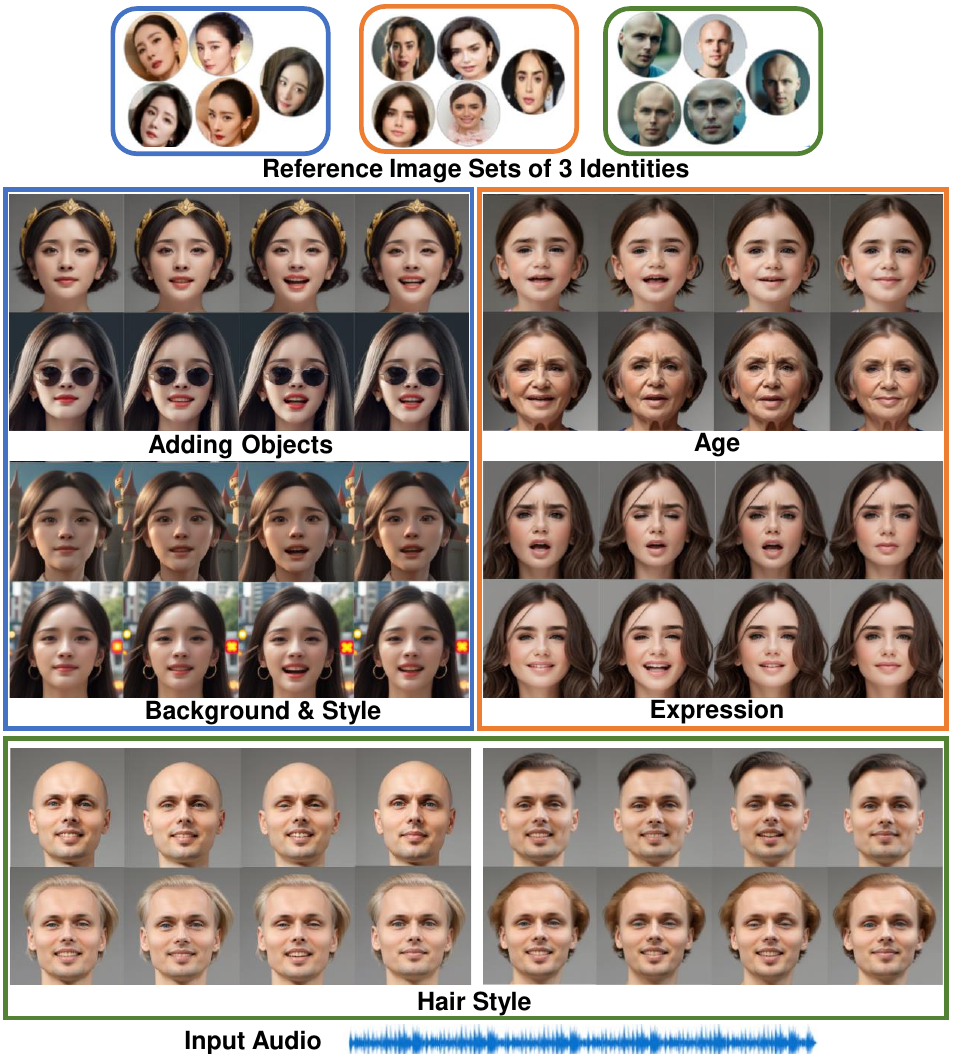}
    \caption{Examples of diverse, expressive, and customizable talking faces generated by PortraitTalk. The model supports a wide range of applications, including identity interpolation, facial expression editing, and the modification of visual attributes such as age, gender, hairstyle, and background. These results demonstrate the effectiveness of PortraitTalk’s multimodal conditioning in enabling fine-grained, user-controllable synthesis while maintaining visual coherence and emotional expressiveness.
    }
    \label{fig4}
\end{figure}

\subsection{Applications and Limitations}
\label{application}
\begin{figure}[!t]
    \centering
    \includegraphics[trim={10.5cm 1.3cm 0cm 1cm}, clip, width=.9\columnwidth]{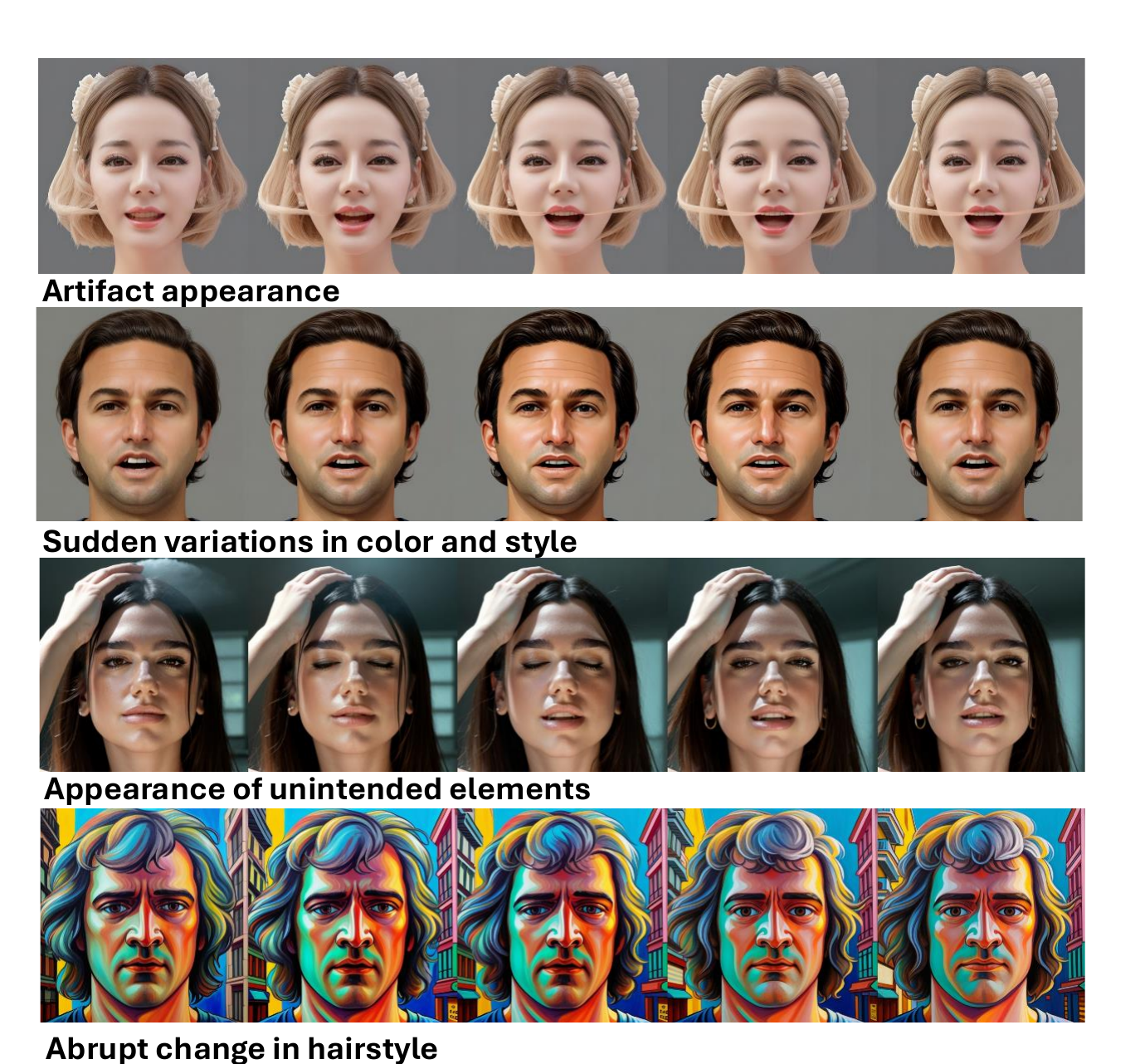}
    \caption{Limitations of the PortraitTalk model. While PortraitTalk supports a wide range of controllable generation features, certain failure cases persist. These include the appearance of unintended elements, changes in color contrast, and alterations in hairstyle that do not align with the intended prompt or reference. Such artifacts are more likely to occur in scenarios involving complex style manipulations or strong emotional expressions.
    }
    \label{fig6}
\end{figure}

PortraitTalk offers extensive customization capabilities for talking face generation through the integration of visual, textual, and audio modalities. 
Fig.~\ref{fig4} and Fig.~\ref{fig0} illustrate a range of applications, including identity interpolation, expression editing, and modifications of style, age, gender, and background. This multimodal conditioning significantly enhances PortraitTalk’s ability to generate emotionally expressive and visually coherent talking faces.
However, certain limitations persist. As shown in Fig.~\ref{fig6}, PortraitTalk occasionally exhibits degraded visual quality and reduced lip synchronization accuracy, particularly when synthesizing intense emotional expressions. In addition, unintended elements, such as hands, may appear during style modifications, likely due to the reliance on CLIP-based text embeddings without dedicated training for complex style alterations. These observations highlight the need for future improvements to achieve a better balance between customization flexibility, expressiveness, and output realism.

\section{Conclusion}
\label{conclusion}
We introduced PortraitTalk, a robust and customizable audio-to-talking face generation framework, comprising IdentityNet and AnimateNet. PortraitTalk preserves the identity and enhances the temporal coherence in video generation. It not only delivers high-fidelity audio-lip synchronization but also offers flexible customization capability via text prompts. This enables the creation of expressive talking faces across varied styles and emotions without retraining. Both quantitative and qualitative assessments confirm the superiority of our model, affirming its effectiveness and practical applicability in real-world scenarios.

\bibliographystyle{IEEEtran}
\bibliography{TIP}

\vfill

\end{document}